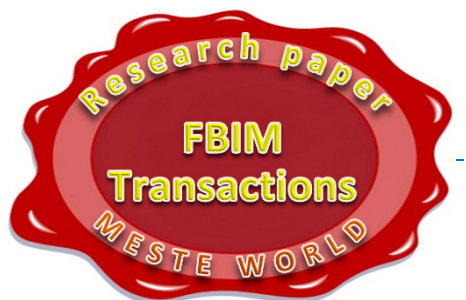

# VEŠTAČKA INTELIGENCIJA U RANGIRANJU VAŽNOSTI 3D ČVOROVA

## ARTIFICIAL INTELLIGENCE ORDERED 3D VERTEX IMPORTANCE

**Iva Vasić**
Faverde SA, Sion, Switzerland

**Bata Vasić**
Poslovni i pravni fakultet, Univerzitet „Union – Nikola Tesla", Beograd, Srbija

**Zorica Nikolić**
Poslovni i pravni fakultet, Univerzitet „Union – Nikola Tesla", Beograd, Srbija

*Apstrakt*

*Rangiranje čvorova višedimenzionalnih mreža je krucijalno u mnogim oblastima istraživanja, uljučujući selektovanje i određivanje važnost odluka. Pojedine odluke imaju značajno veću važnost od ostalih, te je takva i njihova težinska kategorizacija. Ovaj rad definiše novi metod određivanja težine odluka koristeći potpuno novu metodu upotrebe veštačke inteligencije za težinsko rangiranje čvorova trodimenzionalne mreže, unapređujući postojeći Ordered Statistics Vertex Extraction and Tracing Algorithm (OSVETA) baziran na modulaciji kvantizovanih indeksa (QIM) i kodova za ispravljanje grešeka. Tehnika koju predlažemo u ovom radu nudi značajna poboljšanja u efikasnosti odlučivanja o važnosti čvorova mreže u odnosu na statističke OSVETA kriterijume, zamenjujući heuristične metode metodama precizne predikcije modernih neuronskih mreža. Nova tehnika upotrbe veštačke inteligencije omogućava značajno bolju definiciju geomerijske mreže i bolju procenu topologijskih karakteristika. Doprinosi novog metoda rezultuju veću preciznost u definisanju stabilnih čvorova odluke, značajno smanjujući verovatnoću brisanja čvorova odluke.*

**Ključne reči:** Veštačka inteligencija, 3D geometrija, Teorija informacija, Skrivanje informacija.

*Addresa autor*a zaduženog za korespodenciju: **Iva Vasić**   ✉ ivavasic@hotmail.com

*Abstract*

*Ranking vertices of multidimensional networks is crucial in many areas of research, including selecting and determining the importance of decisions. Some decisions are significantly more important than others, and their weight categorization is also imortant. This paper defines a completely new method for determining the weight decisions using artificial intelligence for importance ranking of three-dimensional network vertices, improving the existing Ordered Statistics Vertex Extraction and Tracking Algorithm (OSVETA) based on modulation of quantized indices (QIM) and error correction codes. The technique we propose in this paper offers significant improvements the efficiency of determination the importance of network vertices in relation to statistical OSVETA criteria, replacing heuristic methods with methods of precise prediction of modern neural networks. The new artificial intelligence technique enables a significantly better definition of the 3D meshes and a better assessment of their topological features. The new method contributions result in a greater precision in defining stable vertices, significantly reducing the probability of deleting mesh vertices.*

**Keywords:** *Artificial intelligence, 3D geometry, Information theory, Hiding information*

## 1 UVOD

Svedoci smo revolucije upotrebe veštačke inteligencije i prilagođavanja deep learning tehnika u gotovo svim oblastima. Unapređenja u kompjuterizaciji i brzini Internet komunikacija omogućava napredak ovakvog razmišljanja, pružajući značajni napredak u svim oblastima uključujući bazične težnje u definisanju oblika, ali i širokoj oblasti socioloških i društvenih ponašanja. Interesantno i istovremeno logično je da su se prve revolucionarne ideje upotrebe neuronskih mreža pojavile u oblastima klasifikacije slika (Krizhevsky A., Sutskever I., & Hinton G. E., 2012), (Wu J., Yu Y., Huang C., & Yu K., 2015), i prepoznavanju objekata (Simonyan K. & Zisserman A. 2015). Međutim, uprkos svim idejama poslednje decenije, naučno interesovanje za korišćenje veštačke inteligencije za opis trodimenzionalnih objekata i njihove klasifikacije i prepoznavanja je još uvek u začetku. Razlog ovakvog stanja je u najvećoj meri neuniformna reprezentacija i opis 3D objekata kao i veoma kompleksna analiza geometrijskih mreža, ali i relativno nerazvijeni alati Internet mašinerije za njihovu upotrebu. Ipak, poslednjih nekolio godina se beleži progres u ovom području.

Slično kao i u procesiranju slika, prepoznavanje 3D objekata je počelo istraživanjem 3D oblika i upotrebe neuronskih mreža u te svrhe. Najnovija istraživanja uključuju konvolucione neuronske mreže (CNN) koje postaju aktivne u oblasti prepoznavanja 3D objekata, analizi oblika i sintezi trodimenzionalnih scena (Maturana D. & Scherer S., 2015), (Wang P-S., Liu Y., Guo Y.-X., Sun C.-Y., & Tong X., 2017), (Huang H., Kalogerakis E., Chaudhuri S., Ceylan D., Kim V. & Ymer E., 2017) i (Wang K., Savva M., Chang A. X. & Ritchie D., 2018). Međutim, po našim saznanjima nema rezultata u oblasti vodenih žigova korišćenjem veštačke inteligencije. Prvi objavljen rad iz te oblasti i sasvim nov pristup publikvani su u našem radu koji uključuje više opisnih vektora i obezbeđuje robusnost skrivenih podataka (Vasic B., Raveendran N., & Vasic B., 2019).

U ovom radu razmatramo problem klasifikacije važnih čvorova geometrijske 3D mreže u cilju robusnog i slepog skrivanja podataka unutar mreže. Takođe je izložena matematička formulacija kriterijuma selekcije čvorova sa uključenim algoritmom izračunavanja težinskih koeficijenata neuronske mreže u cilju efikasnog odlučivanja i formiranja vektora indeksa čvorova koji su a priori nepoznati ili zavisni od kompleksnih uslova definisanja oblika okruženja. U odnosu na statistički empirijski model OSVETA algoritma (Vasic B., 2012) novi pristup omogućava ekstrakciju navedenog vektora sa značajnim unapređenjem brzine i tačnosti, što smanjuje verovatnoću brisanja informacije i dalje slobodno korišćenje moćnih kodova za ispravljanje grešaka, a time i tačnije čitanje upisane informacije.

Rad je organizovan na sledeći način. Definicija problema je sumarizovana u Odeljku 2. Odeljak 3 objašnjava relevantna obeležavanja i rezultate OSVETA algoritma, uključujući listu topoloških i geometrijskih kriterijuma karakterizacije geometrijske 3D mreže. U Odeljku 4 prezentujemo teorijske principe konstrukcije neuronske mreže sa osnovnim objašnjnjem backpropagation procesa i optimizacionog procesa učenja neuronske mreže. Na kraju, odeljak 5 prikazuje preliminarne eksperimentalne rezultate performansi verovatnoće brisanja i verovatnoće greške.

## 2 SUMARIZACIJA PROBLEMA

Pardigma nesmetanog skrivanja podataka podrazumeva postizanje pouzdanosti ugradnje i otpornosti upisanih podataka u odnosu na najčešće geometrijske i topološke transformacije, ali i obezbeđivanje adekvatnog algoritma čitanja ugrađenih podataka. Međutim, zbog osnovnog opisa modela suočeni smo sa neuniformnom definicijom 3D modela u prostoru i zbog toga sa sinhronizacionim problemima u procesu čitanja ugrađenih podataka.

Da bismo generizovali sve neophodnosti koje zahtevaju klase robusnih i slepih vodenih žigova moramo definisati važnost primitiva geometrijskih mreža koji će ispunjavati uslove pouzdanih nosioca podataka koje želimo ugraditi. Najpre oni moraju ispunjavati uslove invarijantnosti na osnovne geometrijske i topološke transformacije, ali i veoma desktruktivne i maliciozne algoritme brisanja i reorderinga.

Kako su 3D mreže u svojoj reprezentaciji definisane ogromnim setom podataka, optimizacioni algoritmi često desetkuju broj čvorova minimizujući njihov broj u opisu oblika. Izabrani čvorovi za upis podataka zato moraju biti izabrani tako da učestvuju u opisu oblika i time budu svrstani u grupu važnih primitiva koji imaju veoma malu verovatnoću uklanjanja i reorderinga. Logično, primitivi koji učestvuju u definisanju oblika ispunjavaju sve navedene zahteve. Zbog toga je naš cilj definisanje skupa čvorova koji ispunjavaju uslove pouzdanih nosioca informacije.

## 3 OSVETA

Da bismo definisali neuronsku percepciju oblika počećemo sa diskusijom karakterizacije oblika 3D modela. Međutim, definisanje važnih čvorova koji formiraju jednostavan i prepoznatljiv 3D oblik nije trivijelno. Zbog toga je neophodno najpre definisati problem procene diskretne zakrivljenosti i uočiti razliku u rezultatima određenih metoda.

### 3.1 Definicija algoritma i obeležavanje

OSVETA algoritam se sastoji od tri koraka: i) definisanje i ocenjivanje kriterijuma za procenu važnosti čvorova, ii) precizno praćenje zakrivljenosti oblika i izračunavanje karakterističnih osobina i iii) praćenje važnosti ekstraktovanih čvorova u odnosu na topologiju zadate geometrijske mreže. Preciznije, za geometrijsku mrežu $M(V, F)$, gde su $V$ i $F$ vektori Euklidskih koordinata čvorova i topoloških konekcija između njih respektivno, algoritam kao rezultat daje dva vektora: $\mathbf{s}$, vektor stabilnosti u opadajućem nizu njihovih vrednosti i $\mathbf{q}$, vektor odgovarajućih indeksa. Čvorovi geometrijske mreže, poređani u odnosu na opadajuću stabilnost, formiraju vektor $V_o = \mathbf{v_i}$, i vektor $\mathbf{p}$ dužine $L$, koji je dobijen razmatranjem prvih $L$ elemenata vektora $V_o = \mathbf{v_i}$. Slika 1 ilustruje glavne uglove i oblasti konektovane za razmatrani čvor $v_i$.

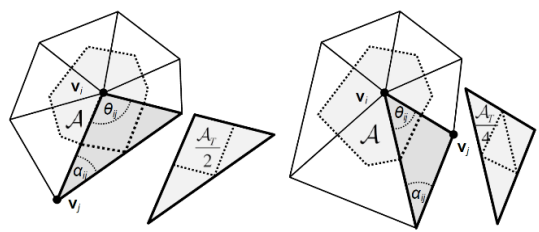

*Slika 1. Ilustracija relevantnih osobina i promenljivih razmatranog čvora $\mathbf{v}_i$*

OSVETA algoritam daje dobre rezultate čak i sa malim brojem deskriptora, ali se i više deskriptora mogu razmatrati u segmentaciji 3D geometrijske mreže (Vasic B., 2012). Da bi ideja bila jasnija, u ovom trenutku mi ćemo pažnju posvetiti samo deskriptorima iz tabele 1 (Vasic B., 2012).

*Tebala 1 OSVETA kriterijumi za procenu važnosti i njihovo rangiranje*

| Br. | Criterijum | Opis | Težina |
| --- | --- | --- | --- |
| 1 | $\psi_{min} \geq 0$ | Pozitivni minimalni dihedralni ugao | 1,0 |
| 2 | $\Theta < 2\pi$ | Mali Teta ugao | 1,0 |
| 3 | $\kappa G1 > 0$ | Pozitivna Gausova zakrivljenost | 1,0 |
| 4 | $\psi_{max} \geq 0$ | Pozitivni maksimalni dihedralni ugao | 0,9 |
| 5 | $\theta > 2\pi$ | Veliki Teta ugao | 0,8 |
| 6 | $\kappa G < 0$ | Negativna Gausova zakrivljenost | 0,8 |
| 7 | $\kappa G1 < 0$ | Negativna Gausova zakrivljenost | 0,7 |
| 8 | $\kappa G > 0$ | Pozitivna Gausova zakrivljenost | 0,4 |

Tabela daje oznake promenljivih, funkcija i uslove za definisane vrednosti karakteristika, ali takođe određuje i težine svih deskriptora koji će kasnije biti korišćene kao inicijalne vrednosti težinskih koeficijenata u backpropagation procesu.

### 3.2 Deskriptori diskretne Gausove zakrivljenosti

Glavni problem u uslovima koji generalizuju izračunavanja geometrijske mreže je tačnost diskretnog izračunavanja mreže. Iz tabele vidimo da je Gausova zakrivljenost jedna od najrelevantnijih deskriptora oblika. Štaviše, ostali deskriptori su implicitno u relaciji sa ovim deskriptorom. Zbog toga i počinjemo kratko razmatranje metoda procene i izračunavanja Gausove zakrivljenosti.

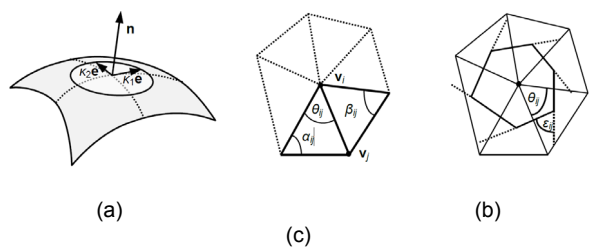

*Slika 2. Ilustracija: (a) beskonačno malo susedstvo površinskog dela; (b) 1-prsten suseda čvora $v_i$ i uglova nasuprot zajedničke ivice; (c) Spoljašnji uglovi Voronoi regiona.*

Iz diferencijalne geometrije (Spivak M., 1999) znamo za *manifold* površinu $M$ u $\mathbb{R}^3$, da se svaka tačka na datoj površini može lokalno aproksimirati tangentnom ravni koja je normalna na vektor *normale* **n**. **K** je definisano kao normalna zakrivljenost krive koja pripada samoj površini i ravni koja sadrži **n** i jedinični vektor pravca **e** u tangetnoj ravni. *Srednju zakrivljenost* $\kappa_H = (\kappa_1 + \kappa_2)/2$ definišemo kao srednju vrednost dve *glavne zakrivljenosti* $\kappa_1$ i $\kappa_2$ površine $M$, Gausova zakrivljenost $\kappa_G$ je definisana kao proizvod dve *glavne zakrivljenosti* $\kappa_G = \kappa_1 \kappa_2$.

Za datu tačku diskretne površine $M$ normala srednje zakrivljenosti (Laplace-Beltrami operator): $\mathrm{K}(\mathbf{v}_i) = 2\kappa_H(\mathbf{v}_i)\mathbf{n}(\mathbf{v}_i)$ daje srednju zakrivljenost $2\kappa_H(\mathbf{v}_i)$ i jedinični vektor normale $\mathbf{n}(\mathbf{v}_i)$ u čvoru $\mathbf{v}_i$. Srednja zakrivljenost i Gausova zakrivljenost diskretne površine koja zavisi samo od pozicije čvora i uglova pripadajućih trouglova respektivno su dati izrazima:

$$\mathbf{K}(\mathbf{v}_i) = \frac{1}{2\mathcal{A}_{Mixed}} \sum_{j \in N_1(i)} (cot\alpha_{ij} + cot\beta_{ij})(\mathbf{v}_j - \mathbf{v}_i)$$

$$\kappa_G(\mathbf{v}_i) = \frac{1}{\mathcal{A}_{Mixed}} \left( 2\pi - \sum_{j=1}^{\#f} \theta_{ij} \right)$$

Gde su #f i $\theta_{ij}$ broj susednih trougaonih površina i ugao $j$-tog susednog trougla u čvoru $\mathbf{v}_i$ respektivno. Polje prvog prstena trouglova oko čvora $\mathbf{v}_i$ je $\mathcal{A}$ (Meyer M., Desbrun M., Schröder P., & Barr A.H., 2003). Tako dobijamo izraz za diskretne zakrivljenosti:

$$\kappa_1(\mathbf{v}_i) = \kappa_H(\mathbf{v}_i) + \sqrt{\Delta(\mathbf{v}_i)}, \quad \kappa_2(\mathbf{v}_i) = \kappa_H(\mathbf{v}_i) - \sqrt{\Delta(\mathbf{v}_i)}$$

gde je:

$$\Delta(\mathbf{v}_i) = \kappa_H^2(\mathbf{v}_i) - \kappa_G(\mathbf{v}_i), \quad \kappa_H(\mathbf{v}_i) = \frac{1}{2}\|\mathbf{K}(\mathbf{v}_i)\|.$$

Sa druge strane ideja procene zakrivljenosti metodom poklapanja kvadrika (Fitting Quadric Curvature Estimation) zasniva se na tome da se glatka geometrijska površina može lokalno aproksimirati kvadratnom polinomskom površinom. Tako metod uklapa kvadrik u lokalno susedstvo za svaku izabranu tačku u lokalnom koordinantnom okviru. Zakrivljenost kvadrika u izabranoj tački je definisan kao procena zakrivljenosti diskretne površine. Za jednostavnu formu kvadrika $z' = ax'^2 + bx'y' + cy'^2$ data je sledeća procedura: i) Procena normale površine **n** u čvoru **v** na jedan od dva načina: jednostavno ili težinsko usrednjavanje ili izračunavanje najmanjeg kvadrata uklopljene površine u čvoru i njegovom okruženju; ii) Pozicioniranje lokalnog koordinantnog sistema $(x', y', z')$ u čvoru **v** i poklapanje $z'$ ose duž procenjene normale. McIvor i Valkenburg (McIvor A.M. & Valkenburg R.J. 1997) sugerišu poklapanje $x'$ koordinantne ose sa projekcijom globalne x ose na tangentnu ravan, koja je definisana pomoću vektora normale **n**. Ovo uzrokuje rotaciju matrice $\mathcal{R}' = [r_1, r_2, r_3]^T$ iz globalnog u lokalni okvir:

$$\mathbf{r}_3 = \mathbf{n} \quad \mathbf{r}_1 = \frac{(\mathcal{I} - \mathbf{n}\mathbf{n}^T)\mathbf{i}}{\|(\mathcal{I} - \mathbf{n}\mathbf{n}^T)\mathbf{i}\|} \quad \mathbf{r}_2 = \mathbf{r}_3 \times \mathbf{r}_1$$

gde je $\mathcal{I}$ identity matrica, a **i** je globalna x osa $[1,0,0]^T$. Ako upotrebimo sugerisana poboljšanja, uklopimo mapirane podatke sa proširenim kvadrikom $\hat{z} = a'\hat{x}^2 + b'\hat{x}\hat{y} + c'\hat{y}^2 + d'\hat{x} + e'\hat{y}$, i rešimo rezultujući sistem linearnih jednačina, možemo izračunati glavne zakrivljenosti $\kappa_1$ i $\kappa_2$.

$$\begin{bmatrix} x_1^2 & x_1 y_1 & y_1^2 \\ \vdots & \vdots & \vdots \\ x_n^2 & x_n y_n & y_n^2 \end{bmatrix} \begin{bmatrix} a' \\ b' \\ c' \end{bmatrix} = \begin{bmatrix} z_1 \\ \vdots \\ z_n \end{bmatrix}$$

Tako dobijamo procenu Gausove i srednje zakrivljenosti.

$$\kappa_1 = a' + c' + \sqrt{(a'-c')^2 + b'^2}$$
$$\kappa_2 = a' + c' - \sqrt{(a'-c')^2 + b'^2}$$
$$\kappa_G = \frac{4a'c' - b'^2}{(1+d'^2+e'^2)^2}, \kappa_H = \frac{a'+c'+a'e'^2+c'd'^2-b'd'e'}{\sqrt{(1+d'^2+e'^2)^3}}$$

### 3.3 Topološke osobine

Druga grupa topoloških osobina koja može biti uzeta u razmatranje je set osobina koje karakterizuju važnost regiona i njegovu otpornost na transformacije. Kako optimizacioni i simplifikacioni procesi destruktivno deluju na geometriju objekta, tako su neki uglovi, čvorovi, ivice, klasifikovani kao riskantni primitivi. Preciznije to su granični čvorovi, čvorovi konektovani kolinearnim uglovima, ali i čvorovi, ivice i površi koje se nalaze u ravnim i glatkim područjima geometrijske mreže. U ovu grupu spadaju i topološke greške (izolovani čvorovi nezavisni u prostoru, čvorovi koji pripadaju samo jednoj graničnoj ivici, kompleksni čvorovi koji spajaju više od dve površine, ukrštene ivice bez zajedničkog čvora). Nasuprot navedenima, neke od topoloških karakteristika su veoma važne za opis oblika (na pr. minimalni i maksimalni dihedralni ugao između dve površine na istoj ivici: $\psi_{max}$ i $\psi_{min}$ respektivno.

Naša nova tehnika opisa 3D oblika koristi neuronsku mrežu (Neural Network - NN) arhitekturu koja uspešno operiše sa osnovnim geometrijskim i topološkim reprezentacijama 3D geometrijskih mreža. U nastavku ćemo objasniti NN arhitekturu sa fokusom na implementaciji OSVETA funkcija.

## 4 ALGORITAM NEURONSKE MREŽE

Nasuprot konvolucionim neuronskim mrežama primenjenim za klasifikaciju 2D slika, Osobina retkosti i greške u izračunavanju 3D konvolucije ozbiljno ograničava trodimenzionalni zapreminski opis geometrijskih 3D mreža i tačaka. Kada se radi o ugradnji podataka, rezultat klasifikacije bi uništio informaciju o čvorovima, pa bi tako uzrokovao i gubitak ugrađene informacije. Naš cilj je definisanje glavnih karakteristika 3D objekta datih njihovom osnovnom reprezentacijom i odgovarajućoj relaciji karakteristika i ulaznog opisa. Prateći principe Neuronskih mreža baziranih na karakteristikama (Feature-based Neural Networks) i uključujući sve relevantne osobine geometrijskih 3D mreža dizajnirali smo FNN koja direktno koristi ulazni vektor svih čvorova kao vrednosti ulaznih neurona neuronske mreže.

### 4.1 Predložena arhitektura neuronske mreže

Matrica ulaznih neurona $I = \{\mathbf{v}, \mathbf{f}\}$ je predstavljena kao skup vektora: i) vektor čvorova $\mathbf{v} = \{\mathbf{v}_i \mid i = 1,....,n\}$, gde je svaki čvor dat svojim vektorom euklidskih koordinata $\mathbf{v}(v_x, v_y, v_z)$ i ii) vektor topološkog kanala $\mathbf{f}(i) = \{f_j \mid j = 1,...,m\}$ sa svim konekcijama čvorova koji formiraju trougaone površine date geometrijske 3D mreže.

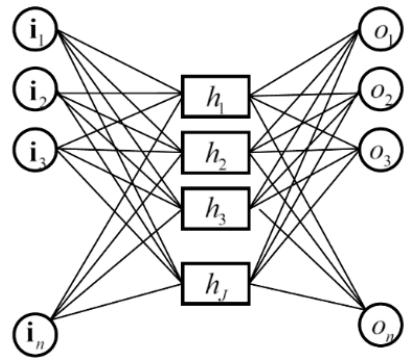

*Slika 3. Arhitektura neuronske mreže: ulazni neuroni predstavljaju elemente vektora i koji sadrži geometrijske i topološke podatke v i f respektivno. Skriveni lejer predstavlja skup funkcija koje opisuju oblik geometrijske 3D mreže (Odeljak 2)*

Na osnovu naših specifičnih zahteva skriveni sloj neurona predstavlja geometrijske i topološke deskriptore i uslove date u Tabeli **1**. Dalje je svaki skriveni neuron povezan kao Fully-Connected Neural Network (FNN) sa izlaznim neuronima koji formiraju izlazni vektor indeksa $\mathbf{o}(i)$. Indeksi su poređani tako da pozicija indeksa $\mathbf{o}(i)$ odgovara važnosti čvora u mreži. Da bi se izbegao problem neregularnosti geometrije i topologije geometrijske 3D mreže, pretpostavili smo da se svaki naš skriveni neuron aktivira jedinstvenom težinskom vrednošću, te da ima takođe jedinstvenu vrednost bias vrednosti. Tako naš algoritam dozvoljava različite dužine ulaznih vektora, a time i različite kompleksnosti ulaznih 3D mreža. Dokazali smo u (Vasic B., 2012) da čak i empirijske vrednosti težinskih koeficijenata obezbeđuju robusnost vodenog žiga ugrađenog u geometrijsku mrežu. Neuronska mreža uči podešavajući sve težinske i bias koeficijente kroz *backpropagation* deo procesa. Tako se na izlazu dobija uređeni vektor indeksa čvorova koji obezbeđuju informaciju o važnosti čvorova. Ova informacija će biti ključna u procesu izbora nosioca za skrivanja informacija. Dalje ćemo objasniti proces učenja neuronske mreže (NN).

## 4.2 Učenje neuronske mreže u backpropagation procesu

Da bismo ilustrovali proces učenja u backpropagation procesu razmatraćemo skriveni neuron $H$ koji prima ulazne signale od ulaznih neurona $I$. Aktivacija tih neurona su zapravo komponente ulaznog vektora $I = \{\mathbf{i}_i \mid i = 1, \ldots, n\}$ respektivno. Obeležimo težinske koeficijente između ulaza $I_I$ i skrivenih neurona $H_J$ sa $\omega_{IJ}$. Na primer, mrežni ulaz $hin_j$ ka $j$-tom neuronu $H$ je zbir težina od neurona $\mathbf{i}_1, \mathbf{i}_2, \ldots, \mathbf{i}_n$: $hin_j = b_j + \sum_i \mathbf{i}_i \omega_{ji}$. Bias $b$ je uključen dodavanjem ulaznom vektoru i tretiran je identično kao svaki drugi težinski koeficijent. Aktivacija $h$ skrivenog neurona $H$ zadat je nekom funkcijom njegovog mrežnog ulaza $h = g(hin)$, gde je sigma funkcija $\sigma(\mathbf{i}) = (1 + e^{-\mathbf{i}})^{-1}$. Indeksi $JK$ su analogijom korišćeni za obeležavanje težinskih koeficijenata $\omega_{JK}$ između skrivene jedinice $H_J$ i izlazne jedinice $O_K$, koje su proizvoljne ali fiksne vrednosti. Na osnovu ovakvog označavanja, odgovarajuća mala slova mogu označavati zbirne indekse u izvodu uslova nadogradnje težina. Za proizvoljnu aktivacionu funkciju $g(x)$ njen izvod je označen kao $g'$. Zavisnost aktivacije od rezultata težinskog koeficijenta usled primene aktivacione funkcije $g$ na mrežni ulaz, da bismo odredili $g(o_K)$, data je kao $o_K = \sum_j h_j \omega_{jK}$.

Težinski koeficijenti se nadograđuju na način prikazan u 4.2.1 i 4.2.2.

### 4.2.1 Nadogradnja težinskih koeficijenata skrivenog lejera

Neka je $\mathbf{t} = (t_1, t_2, \ldots, t_n)$ trening ili ciljni izlazni vektor. Tada je greška koju moramo minimizirati $E = (1/2) \sum_k [t_k - o_k]^2$. Koristeći lančani uslov (Fausett L.V., 1994), imamo:

$\sigma_K = \partial E / \partial \omega_{JK} = [t_K - o_K] g'(oin_k)$. Za težinske koeficijente u konekciji sa skrivenim slojem imamo $H_J$: $\partial E / \partial \omega_{IJ} = -\sum \sigma_k \omega_{Jk} g'(hin_J)[\mathbf{i}_I]$, gde je sigma funkcija data sa $\sigma_J = \sum_k \sigma_k \omega_{Jk} g'(hin_J)$. Gradijent težinskih koeficijenata ka skrivenim neuronima dat je kao $\Delta \omega_{ij} = \eta \sigma_j \mathbf{i}_i$. Kako je svaki naš neuron skrivenog lejera, od ulaznih neurona rangiran istim težinskim koeficijentom i bias-om, gradijent težinskih koeficijenata je dat kao $\Delta \omega_j = \eta \sigma_j \mathbf{i}$.

### 4.2.2 Nadogradnja težinskih koeficijenata izlaznog lejera

Za izračunavanje izlaznih težina koristićemo standardnu FNN šemu koja minimizuje zbir kvadrata greške za sve izlazne neurone: $E = (1/2) \sum_k (h_l - o_k)^2$. Dalje, potrebno je odrediti smer u kome će se menjati težinski koeficijenti. Izračunavamo negativni gradijent greške $\nabla E$ u odnosu na težine $\omega_{kj}$ i tako podešavamo vrednosti težina. Pretpostavimo da je aktivacija tog čvora jednak narednom mrežnom ulazu, tada je izlaz $j$-tog neurona $iin_j = g_j(hin_j)$, pa je tako aktivacija izlaznih neurona $oin_k = \sum_j \omega_{kj} iin_j + b_k$, gde je naša aktivacija izlaza $o_k = g_k(oin_k)$. Kada govorimo o promeni težina, koristićemo ih tako da budu proporcionalne negativnom gradijentu:

$$\partial E / \partial \omega_{kj} = (h_k - o_k) g'_k(hin_k) oin_j.$$

Neka je $\eta$ brzina učenja NN koja je korišćena za kontrolu veličine podešavanja težine u svakom koraku treninga neuronske mreže. Tada se težinski koeficijenti izlaznog sloja nadograđuju prema: $\omega_{kj}(t+1) = \omega_{kj}(t) + \Delta \omega_{kj}(t)$, pri čemu je, $\Delta \omega_{kj} = \eta (h_k - o_k) g'_k(hin_k) iin_j$, dok se nadogradnja težine sigma funkcijom obavlja prema izrazu: $\omega_{kj}(t+1) = \omega_{kj}(t) + \eta \sigma_k iin_j$, gde je $\sigma_k = (h_k - o_k) g'_k(oin_k)$.

### 4.3 Trening ulaz

Na osnovu našeg ulaznog pravila možemo koristiti bilo koji trodimenzionalni vektor bez obzira na broj njegovih elemenata. Istovremeno, dimenzija podataka koje ugrađujemo u 3D geometriju određuje broj neophodnih nosećih čvorova, tako da naša neuronska mreža može učiti koristeći trening vektor čija je dimenzija jednaka ili veća od broja bita koji će biti ugrađeni. Sa druge strane, naš algoritam za skrivanje podataka (Vasic B. & Vasic B., 2013) dostiže kapacitete vodenih žigova koji prevazilaze navedeni problem.

## 5 REZULTATI

Kao što smo pomenuli, za sada ne postoje algoritmi za upotrebu NN kod ugradnje podataka niti adekvatne baze podataka sa 3D modelima koje se mogu koristiti za segmentaciju i prepoznavanje 3D modela. Zbog toga se ovaj rad oslanja uglavnom na teoretskim principima i eksperimentalnim izračunavanjima korišćenjem skupa podataka koji se baziraju na OSVETA algoritmu i njegovoj preciznosti, podjednakoj u izračunavanju karakteristika 3D modela jednostavnih i kompleksnih geometrijskih i topoloških struktura. Naš skup podataka je izveden iz 3D modela korišćenjem funkcija optimizacije i simplifikacije, kao i upotrebom Gausovih filtara koji simuliraju greške u 3D geometriji. Na osnovu jedinstvenih zahteva, rezultujući trening skup je konstruisan korišćenjem algoritma evaluacije i semantičkih korelacija sa geometrijskim i topološkim karakteristikama.

### 5.1 Karakteristične funkcije skrivenog lejera

Definicija oblika geometrijske 3D mreže nije trivijalna i može se pokazati da mnoge karakteristike imaju uticaj na njenu karakterizaciju. Međutim, neke osobine imaju veći uticaj na opis oblika modela ali i perceptualni doživljaj geometrije. Na primer, selekcija važnih čvorova geometriske 3D mreže u zavisnosti od pozitivne vrednosti Gausove zakrivljenosti $\kappa_G > 0$ korišćenjem dva različita kriterijuma, opisana u Odeljku 3.2, prikazana je na Slici 4.

Može se primetiti da se čvorovi selektovani ovim metodama nalaze na potpuno različitim

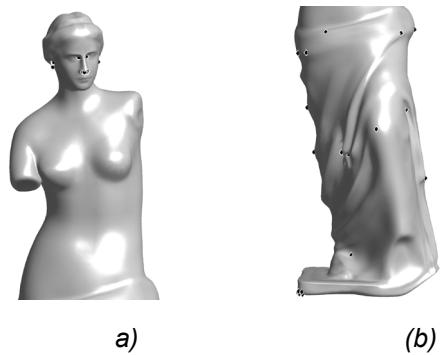

*a)         (b)*

*Slika 4. 3D pozicija prvih 50 čvorova selektovanih kriterijumima $\kappa_G$ i $\kappa_{G1}$ : (a) κG>0- kružići, (b), κG1>0 - romboidi.*

regionima 3D modela iako su svi selektovani čvorovi veoma važni u opisu oblika geometrije. Može se pokazati takođe da ostali kriterijumi dati u Tabeli 1 selektuju različite važne regione i čvorove koji se mogu upotrebiti kao nosioci skrivene informacije. Štaviše, neki od kriterijuma, prikazani u (Vasić, 2012) mogu se takođe koristiti za uklanjanje nevažnih čvorova i regiona i tako smanjiti broj ulaznih i izlaznih neurona, što dovodi do povećanja brzine i efikasnosti procesa učenja naše neuronske mreže.

### 5.2 Verovatnoća brisanja čvorova

Rezultat invarijantnosti čvorova u odnosu na optimizacioni proces dobijen je korišćenjem Neuronske verzije OSVETA i bazičnog OSVETA algoritma. Za simuliranje kanala brisanja korišćen je 'Pro Optimizer' modifikator iz 3D Studio Max 2015 paketa (Kauffman M., 2009) koji se bazira na simplifikacionom perceptualnog oštećenja modela i time do njegove neupotrebljivosti.

Rezultat invarijantnosti čvorova u odnosu na optimizacioni proces dobijen je korišćenjem Neuronske verzije OSVETA i bazičnog OSVETA algoritma. Za simuliranje kanala brisanja korišćen je 'Pro Optimizer' modifikator iz 3D Studio Max 2015 paketa (Kauffman M., 2009) koji se bazira na simplifikacionom algoritmu i decimaciji čvorova. Čak i sa relativno malim trening skupom za našu NN, eksperimentalni test stabilnosti 1000 čvorova, odabranih upotrebom novog algoritma pokazuje superiornost u odnosu na proizvoljno odabran skup istog broja čvorova, ali i značajno unapređenje rezultata u odnosu na naš osnovni algoritam. Rezultati su sumirani u Tabeli 2.

*Tabela 2 Broj čvorova 3D modela uklonjenih procesom optimizacije*

|  | 0% | 20% | 40% | 60% | 80% | 90% |
|---|---|---|---|---|---|---|
| Total VR | 17350 | 12209 | 6953 | 3926 | 2315 | 1448 |
| Random | 0 | 332 | 622 | 781 | 872 | 920 |
| OSVETA | 0 | 1 | 30 | 147 | 332 | 522 |
| Neuro-OSVETA | 0 | 0 | 22 | 121 | 282 | 421 |

Ukupan broj preostalih čvorova nakon brisanja simplifikacijom modela (Total RV) sa datim procentom izbrisanih čvorova, dat je u prvom redu tabele. U poređenju sa brisanjem proizvoljno selektovanih čvorova (drugi red), treći i četvrti red pokazuju broj izbrisanih čvorova od 1000 selektovanih OSVETA i Neuro-OSVETA algoritmom respektivno. Jasno se može primetiti superiornost naših algoritama pri manje akresivnoj optimizaciji. Važno je napomenuti da agresivna simplifikacija dovodi do drastičnog perceptualnog oštećenja modela i time i do njegove neupotrebljivosti.

## 6  ZAKLJUČAK

Ovaj rad prezentuje novu sistematičnu tehniku upotrebe veštačke inteligencije u odabiru važnih primitiva geometrijske 3D mreže, zamenjujući statističke metode pristupa OSVETA algoritma jedinstvenom arhitekturom neuronske mreže, prilagođene specifičnim zahtevima karakterizacije trodimenzionalnih modela. Novi algoritam omogućava precizniju i bržu procenu geometrijske i topološke strukture objekata boljom procenom zakrivljenosti regiona mreže, te tako obezbeđuje selekciju čvorova koji najviše učestvuju u definisanju oblika modela. Ovako izabrani skup se dalje može koristiti za nesmetano skrivanje informacija upotrebom raznih metoda vodenih žigova koji u svojim zahtevima podrazumevaju robusnost skupa nosioca informacije. Iz naših eksperimentalnih rezultata primećujemo značajnu redukciju verovatnoće brisanja nosioca informacije, selektovanih novim algoritmom.

Naše dalje interesovanje će se bazirati na optimizaciji NN parametara i obezbeđenju većeg trening seta 3D modela, koji će unaprediti preciznost odluke neuronske mreže.

## CITIRANA DELA

*Kako citirati ovaj rad? / How to cite this article?*

*Style – **APA** Sixth Edition:*

Vasić, I., Vasić, B., & Nikolić, Z. (2020, 10 15). Veštačka inteligencija u rangiranju važnosti 3D čvorova. (Z. Čekerevac, Ur.) *FBIM Transactions, 8*(2), 193-201. doi:10.12709/fbim.08.08.02.21

*Style – **Chicago** Sixteenth Edition:*

Vasić Iva, Vasić Bata i Nikolić Zorica. 2020. „Veštačka inteligencija u rangiranju važnosti 3D čvorova." Urednik Zoran Čekerevac. *FBIM Transactions* (MESTE) 8 (2): 193-201. doi:10.12709/fbim.08.08.02.21.

*Style – **GOST** Name Sort:*

**Vasić Iva, Vasić Bata i Nikolić Zorica** Veštačka inteligencija u rangiranju važnosti 3D čvorova [Časopis] // FBIM Transactions / ur. Čekerevac Zoran. - Beograd : MESTE, 15 10 2020. - 2 : T. 8. - str. 193-201.

*Style – **Harvard** Anglia:*

Vasić, I., Vasić, B., & Nikolić, Z., 2020. Veštačka inteligencija u rangiranju važnosti 3D čvorova. *FBIM Transactions,* 15 10, 8(2), pp. 193-201.

*Style – **ISO 690** Numerical Reference:*

Veštačka inteligencija u rangiranju važnosti 3D čvorova. ***Vasić, Iva, Vasić, Bata i Nikolić Zorica.** [ur.]* Zoran Čekerevac. 2, Beograd : MESTE, 15 10 2020, FBIM Transactions, T. 8, str. 193-201.